\documentclass[conference]{IEEEtran}
\IEEEoverridecommandlockouts
\usepackage{cite}
\usepackage{amsmath,amssymb,amsfonts}
\usepackage{algorithmic}
\usepackage{graphicx}
\usepackage{acronym}
\usepackage{url}
\usepackage{balance}
\usepackage[hidelinks]{hyperref}
\usepackage{textcomp}
\usepackage{eurosym}
\usepackage{siunitx}
\usepackage{xcolor}
\usepackage{comment}
\usepackage{booktabs}
\usepackage{todonotes}
\usepackage{colortbl}
\usepackage{makecell}
\usepackage{float} 
\usepackage{censor}
\def\BibTeX{{\rm B\kern-.05em{\sc i\kern-.025em b}\kern-.08em
    T\kern-.1667em\lower.7ex\hbox{E}\kern-.125emX}}
\hypersetup{colorlinks=true,citecolor=black, linkcolor=black, urlcolor=blue}
\usepackage{tikz}
\newcommand\copyrighttext{%
  \footnotesize © 2025 IEEE. Personal use of this material is permitted. Permission from IEEE must be obtained for all other uses, in any current or future media, including reprinting/republishing this material for advertising or promotional purposes, creating new collective works, for resale or redistribution to servers or lists, or reuse of any copyrighted component of this work in other works.}
\newcommand\copyrightnotice{%
\begin{tikzpicture}[remember picture,overlay]
  \node[anchor=south,yshift=10pt] at (current page.south) {\fbox{\parbox{\dimexpr\textwidth-\fboxsep-\fboxrule\relax}{\copyrighttext}}};
\end{tikzpicture}%
}
\begin{document}
\acrodef{BEV}{Bird's-Eye View}
\acrodef{CNN}{Convolutional Neural Network}
\acrodef{CV}{Connected Vehicle}
\acrodef{FoV}{Field of View}
\acrodef{FPS}{Frames Per Second}
\acrodef{SIN}{Situation Identification Network}
\acrodef{V2X}{vehicle-to-everything}

\title{Situation Awareness for Intelligent Data Distribution in Connected Vehicles}

\makeatletter
\newcommand{\linebreakand}{%
  \end{@IEEEauthorhalign}
  \hfill\mbox{}\par
  \mbox{}\hfill\begin{@IEEEauthorhalign}
}
\makeatother

\author{
\IEEEauthorblockN{{Falk Dettinger, Akshay Narla and Michael Weyrich}}
\thanks{The first two authors contributed equally to this publication.}
\thanks{The authors would like to thank the German Federal Ministry of Education and Research (BMBF) (under Grant Number: 16MEE0472) and the Chips Joint Undertaking for the financial support under Grant Agreement No: 101139789 (HAL4SDV). The responsibility for the content of this publication lies with the authors.}
\IEEEauthorblockA{
\textit{Institute of Industrial Automation and Software Engineering (IAS)} \\
\textit{University of Stuttgart} \\
{Pfaffenwaldring 47, 70550 Stuttgart, Germany} \\
{E-Mail: \{falk.dettinger, akshay.narla, michael.weyrich\}@ias.uni-stuttgart.de}}
}


\maketitle
\copyrightnotice

\begin{abstract}
The limitations of on-board sensors and blind spots caused by occlusion cause the reduction of perception quality in autonomous vehicles. In such cases, cooperative perception provides additional data via Vehicle-to-Everything communication to enhance local perception, causing a large volume of data transmission. The vehicle can focus on acquiring and utilizing relevant data according to the prevailing road context by identifying the current traffic situation. To achieve this, we propose a concept for the situation identification of the vehicle using Bird's-Eye-View images. Firstly, the situation around the vehicle is identified using object detection with semantic segmentation, followed by understanding the context of the traffic using a situation identification module consisting of an open-source projective transformation network Cam2BEV and a situation identification neural network. The concept was evaluated and validated by running the software on the CARLA simulator using the in-built RGB camera and the semantic segmentation camera. Additionally, the portability of the situation identification module for real-world applications was verified on Cityscapes and nuScenes urban driving datasets. Overall, the proposed situation identification approach enables efficient sensor data management by prioritizing relevant data to the current traffic situation. The source code is available in the following link: \url{https://github.com/akshaynarla/DySi_Select}
\end{abstract}

\section{Introduction}
Autonomous driving is seen as an opportunity to significantly improve the efficiency and safety of transportation by using a comprehensive view of the environment \cite{8911694}. However, despite many sensors, powerful computing systems, developed algorithms, and support for artificial intelligence, no fully autonomous vehicle has yet been approved for SAE Level 5. Therefore, attempts are being made to extend vehicle functions to connected vehicle functions using communication technologies such as 5G \cite{Dettinger2024}. These use the communication capabilities of the connected vehicle (CV) to exchange data and information with other vehicles and back-end systems, which can improve the overall performance of the functions.

Broadcasting and groupcasting approaches are often used to distribute data and information in vehicular environments \cite{10097602}. As the number of connected functions and vehicles increases, the redundancy of messages in the network and, therefore, the required bandwidth will also increase. This is particularly challenging in high-density traffic environments, as redundant transmission places a heavy load on the communication medium \cite{Dettinger_V2X2024}. This reduces the bandwidth per subscriber and increases the transmission time. Therefore, approaches are needed that specifically optimize the transmission of data and information.

Additionally, current state-of-the-art perception methods only identify objects in 3D and do not directly identify the traffic situation. The vehicle has to rely on identifying traffic signs and signals along with normal object detection and understand the situation by relating the various identified objects, which requires a lot of computational resources \cite{DBLP:journals/corr/abs-2106-11342}. In this context, the current work proposes a \textit{Situation Identification neural Network}, which identifies the traffic situation directly based on the top-down projection of camera images, also known as bird's-eye view, of the ego-vehicle as input.

The rest of the paper is organized as follows: section \ref{sec:related_work} outlines the related work, while section \ref{sec:concept} describes our proposed system for situation identification. A detailed evaluation is presented and discussed in section \ref{sec:evaluation} along with the evaluation setup. Finally, in section \ref{sec:conclusion}, we summarize our main findings.  

\section{Related Work}\label{sec:related_work}
Endsley \cite{endsley1995toward} defines situation awareness in three levels: Level 1 describes the recognition of objects and elements in the immediate environment; Level 2 focuses on interpreting the current situation based on those objects; and Level 3 is about predicting future states. This approach can be directly applied to situation awareness in autonomous vehicles, where the levels are represented by environmental perception, situation detection, and situation prediction \cite{ignatious2023analyzing}. Recently, various approaches have been presented to enable context-aware scene understanding in autonomous vehicles, aligning with the approach outlined by Endsley \cite{endsley1995toward}.

In \cite{yu2023scene}, Yu et al. propose an approach for road scene understanding and risk assessment of driving maneuvers based on spatio-temporal scene-graphs. In their approach, the authors use a Faster RCNN and OpenCV's perspective transformation library to detect objects within the cameras FoV, respectively, to generate a birds-eye-view (BEV) image of the environment. The authors model the situation in a graph to derive insights but do not explicitly identify it.

The authors in \cite{10217340} suggest a scene understanding framework for RGB-and-thermal images to compensate for the degradation of RGB images in bad lighting conditions using a Dynamic Bilateral Cross-Fusion (DBCNet) Module. The results on the \textit{PST900 Dataset} show that thermal images improve scene understanding. Nevertheless, the situation itself is not identified by this method.

As trajectory planning in autonomous driving depends on the behavior of other participants, understanding driving scenarios is crucial. A motion learning and prediction approach is proposed in \cite{9921850} to tackle this. The proposed CASPNet consists of an architecture with several neural networks and attention blocks. The evaluation shows state-of-the-art results on the nuScenes prediction challenge. However, the context of the surrounding situation is not immediately identified.


Reiher et al. \cite{ReiherLampe2020Cam2BEV} provides a methodology, \textit{Cam2BEV}, for transforming the camera images from front plane to BEV plane. A spatial transformer based neural network, uNetXST, performs projective transformations and converts the image to BEV plane. Although this method identifies the objects around the ego-vehicle in BEV, the overall context of the traffic situation is not identified. Nevertheless, our work builds upon the transformation provided by this \textit{Cam2BEV} network to help in identifying the situation. 

Section \ref{sec:related_work} highlights that context detection and situation awareness are necessary components in Advanced Driving Assistant Systems (ADAS) and autonomous vehicles, enabling improved environmental detection and interaction. Given the dynamic and diverse nature of the road scenarios, the reviewed articles focus on machine learning-based approaches to improve aspects such as object detection, scene understanding, or trajectory prediction in ADAS systems without considering the explicit road situation of the vehicle. In this article, we explore using a machine learning-based approach to predict and interpret road traffic situations. To our knowledge, no other methods or publications provide a concept for directly identifying the traffic situation from ego-vehicle's BEV. 

\section{Concept}\label{sec:concept}
In this section, we introduce the concept developed to identify the situation. To achieve this, as defined by Endsley \cite{endsley1995toward}, it is necessary to identify the objects in the environment initially before understanding the overall context. We use semantic segmentation for object detection followed by transformation of the camera view to BEV to better understand the situation in the vicinity of the ego-vehicle. The overview of the concept is visualized in Fig.~\ref{SysArch}, and details of the functional modules of the concept are explained in Section~\ref{subsec:semseg}, \ref{subsec:cambev} and \ref{subsec:sin} respectively.
\begin{figure*}[btp]
    \centering
    \includegraphics[scale=0.35]{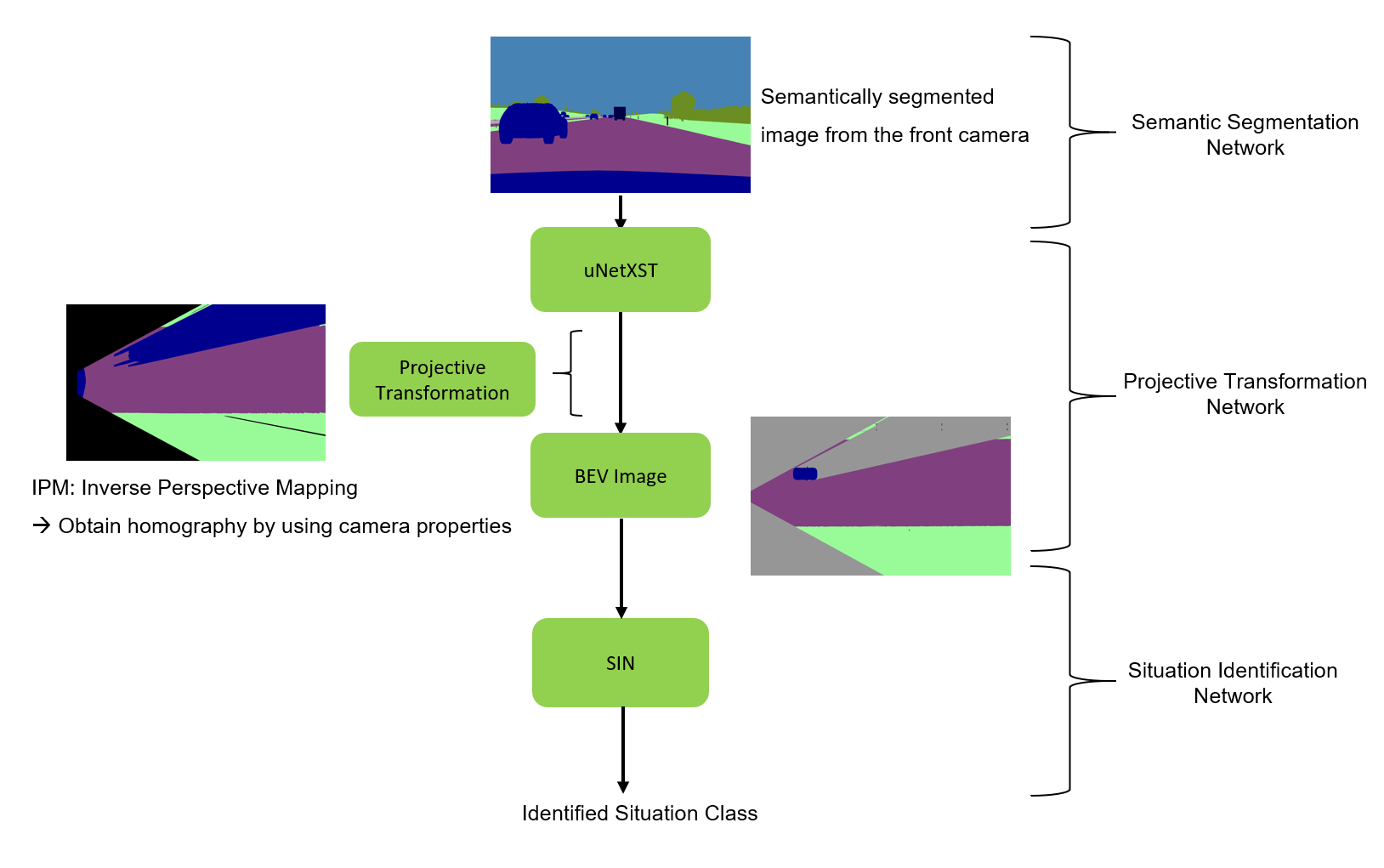}
    \vspace{-10pt}
    \caption{Proposed concept for traffic situation identification}
    \label{SysArch}
\end{figure*}

\subsection{Semantic Segmentation Network (SemSeg)}\label{subsec:semseg}
The \textit{SemSeg} module implements the semantic segmentation for the input RGB image frames from the front camera. An off-the-shelf semantic segmentation algorithm, DeepLabv3+ \cite{chen2018encoderdecoder}, is used to identify the objects around and produce a semantically segmented image. This acts as the input to the \textit{Cam2BEV} module.

\subsection{Projective Transformation Network (Cam2BEV)}\label{subsec:cambev}
\textit{Cam2BEV} neural network, introduced in \cite{ReiherLampe2020Cam2BEV}, forms the backbone of the concept and converts the semantically segmented image frames to obtain a corrected BEV image with a special semantic class for occlusions. \textit{Cam2BEV} realizes this by using a spatial transformer unit that projectively transforms the extracted feature maps from the input using a \textit{uNetXST} network. This allows direct translation of the semantically segmented images from the front to birds-eye view. 

\subsection{Situation Identification Network (SIN)}\label{subsec:sin}
The Situation Identification neural Network, or \textit{SIN}, is the most important contribution of the work and will be used to identify the situation of the ego-vehicle based on the BEV image from the projective transformation module. The context of the road situation is more easily understood in the BEV than in the front view since it provides clear and comprehensive information about the surroundings. The BEV output from \textit{Cam2BEV} acts as the input to the \textit{SIN} to identify one of the defined situation classes introduced. This module's functional scope and task is to provide one of the five configured situation classes based on the trained neural network. For this purpose, a separate new dataset, namely the \textit{SIN} dataset, is derived from the synthetic \textit{Cam2BEV} dataset. 

\subsubsection{Situation Classes} \label{subsec:situationclasses}
Five situation classes have been defined based on normal on-road driving and the possible on-road occlusion scenarios. These are also some of the most common traffic situations possible on most roads other than highways where V2X becomes relevant. The situation classes and their definitions in the context of this work are as follows:
\begin{itemize}
    \item Free Intersection: Ego-vehicle is the lead vehicle while reaching an intersection.
    \item Occluded Intersection: Ego-vehicle is occluded by a vehicle in front, limiting the overall FoV at an intersection.
    \item Free Drive: Ego-vehicle has no occlusions on road.
    \item Free Drive with Parked Vehicles: Ego-vehicle has no occlusions from other traffic vehicles, but from parked vehicles or buildings.
    \item Occluded Drive: Ego-vehicle is occluded by a moving vehicle in the driving direction, limiting the overall FoV.
\end{itemize}

\subsubsection{Dataset} To use the \textit{SIN} module, training of the network in a custom dataset is necessary. This dataset, called the \textit{SIN} dataset, is derived from the open-source \textit{Cam2BEV} dataset and consists of 2500 training images, 250 validation images, and 100 testing images at a resolution of $1936\times968$. These BEV images are divided equally into the aforementioned situation classes. The newly derived dataset trains the neural network model to identify the situation. Currently, the dataset is only created for the BEV of the front camera images.


\subsubsection{Neural Network Architecture}
Since the new \textit{SIN} dataset is relatively small and consists of images with similar backgrounds and minimal foreground data, using a complex neural network or the state-of-the-art method is not necessary. A simple CNN, VGGNet-16\cite{simonyan2015deep}, is selected as the backbone for the Situation Identification Network to learn the features from an already semantically segmented input. The architecture has been successfully applied in various classification applications using a diverse range of datasets \cite{10212250, https://doi.org/10.1002/cpe.6767       }.

The identified situation can then be used to make rule-based decisions in other applications. This can make the performance of the software system in the vehicle more robust and reliable. Since the proposed concept uses semantic segmentation to process information, it can be assumed that the realized system would also be suitable for real-world data, which will be evaluated in section \ref{sec:evaluation}.


\section{Evaluation}\label{sec:evaluation}  
\subsection{Model Training and Performance}\label{model} 
The system consists of three cascaded neural networks, with each network model trained separately on different datasets to perform according to the application's needs. The performance of the open-source and the out-of-the-box models are not discussed in detail, as they can be obtained in the respective references.

\subsubsection{SemSeg module} 
The out-of-the-box semantic segmentation is a DeepLabv3+ network \cite{chen2018encoderdecoder} pre-trained on the Cityscapes dataset. No application-specific training was performed on the \textit{SemSeg} module. Since the network was used with pre-trained weights, the performance is not considered for evaluating the \textit{SemSeg} module.

\subsubsection{Cam2BEV module} 
The \textit{Cam2BEV} module is trained with the 2F variant from the open-source \textit{Cam2BEV} dataset. This dataset, created using simulation tools, provides ground truth BEV images with an approximate field of view of 50 m x 25 m. The semantically segmented images with Cityscapes color palette are reduced to ten classes using one-hot encoding with the introduction of the "occluded" class. The loss function for training these ten classes is retained as provided in the open-source code. The \textit{uNetXST} neural network backbone is trained for 80 epochs using Adam optimizer with a learning rate of 0.0001 applied to batches of size 16.

\subsubsection{SIN module} 
The \textit{SIN} module is trained with the newly derived training dataset containing the ground truth BEV images of each situation class using transfer learning. The non-trainable layers of the VGGNet-16 architecture are initialized with imagenet weights. The trainable weights are modified during the training on the new dataset. This network architecture is trained for 25 epochs using Adam optimizer with a learning rate 0.0001 applied to batches of size 15. The model is trained for 25 epochs only since the dataset is small to prevent over-fitting.

\subsection{Evaluation setup}
Although the newly derived dataset contains traffic situation classes that are possible on any roads, i.e., urban, rural, highways, or other roads, the concept is validated in an urban environment using the CARLA simulator and with real-world urban scenario datasets nuScenes\cite{7780719} and Cityscapes\cite{9156412} as shown in Section \ref{subsec:results}. In the following subsection, the evaluation setup used for evaluating the concept in simulation and real-world data is explained in detail. 

The training, simulation, and evaluation were made on the following hardware:
\begin{itemize}
    \item CPU: Intel Xeon E5-2695 V4
    \item GPU: 4$\times$ NVidia Tesla K80 11 GB graphic memory each
    \item RAM: 512 GB DDR4 2400 MHz
\end{itemize}

\subsubsection{Simulation}\label{subsec:simu} 
The concept is validated using the CARLA simulation framework, which provides the necessary environment for simulating vehicles and traffic on a pre-defined urban town map. The goal of the simulation is to demonstrate that the traffic situation of the ego-vehicle is accurately identified.

The simulation is controlled via a Python script, which manages the CARLA world and allows dynamic software changes. The CARLA server provides the necessary urban environment and synchronizes all the clients involved in the simulation. The ego-vehicle component referred to as the DySi\_Select client spawns the ego-vehicle with the camera sensor and allows the software to run based on the neural network pipeline explained in Section~\ref{sec:concept}. Here, two camera settings are tested since CARLA offers a semantic segmentation camera with Cityscapes color palette and a regular RGB camera. All the other traffic vehicles are programmed in DySi\_Select Traffic python client to drive at an optimal velocity suitable for urban conditions. Additionally, parked vehicles are also spawned to increase possible occluded scenarios. We run the ego-vehicle in the CARLA map of Towns 01, 02, and 03 for 30 min in varied traffic density and road conditions.

\subsubsection{Real-world Portability}\label{subsec:realworld} 
The developed concept is evaluated using the images from nuScenes\cite{7780719} and Cityscapes\cite{9156412} to validate the software pipeline used in the simulation prototype for real-world applications. The network models are initialized with pre-trained weights per Section~\ref{model}. The frames from the dataset are processed sequentially through a series of neural networks. The situation is predicted every 10th frame to avoid redundant processing, assuming a consistent vehicle condition. This approach is effective because the situation rarely changes from one frame to the next if we assume 30 FPS as the normal camera frame rate for ADAS. By verifying the performance of the developed software on these datasets, we ensure that the concept is portable for use in actual vehicles.

\subsection{Results and Discussion}\label{subsec:results}
In this section, we discuss the evaluation results based on the evaluation setups explained in the previous section. The performance of the software is evaluated by considering the accuracy and the average inference time for situation prediction in both the simulation prototype and the real-world data. The correct situation inference is verified manually by cross-verifying the predicted situation with the ground truth images from the ego-vehicle camera.

\subsubsection{Simulation} 
Using the high-resolution semantic segmentation camera provided by CARLA, the software (without the \textit{SemSeg} module) was evaluated by simulating as mentioned in \ref{subsec:simu}. It can be observed from Table~\ref{tab:semsegperf} that the rate of correct situation inference ranges from 67.3\% in Town 03 to 80.2\% in Town 02. The average inference time of the situation here is at 0.26s, with semantically segmented images as the input to the pipeline.

\begin{table}[htp]
\caption{Performance of the CARLA prototype with semantic segmentation camera}
\label{tab:semsegperf}
\centering
\begin{tabular}{|c|c|c|c|} \hline  
 
 Town& Processed Images &Correct Inference &Accuracy\\ \hline  
 Town 01&  636& 481&75.6\%\\ \hline  
 Town 02&  1948& 1563&80.2\%\\ \hline  
 
 Town 03& 700& 471&67.3\%\\ \hline 
 Overall& 3607& 2768&76.7\%\\\hline
\end{tabular}
\end{table}

\begin{table}[htp]
\caption{Performance of the CARLA prototype with RGB camera}
\label{tab:rgbperf}
\centering
\begin{tabular}{|c|c|c|c|} \hline  
 Town& Processed Images &Correct Inference &Accuracy\\ \hline  
 Town 01&  207& 136&65.7\%\\ \hline  
 Town 02&  195& 124&63.6\%\\ \hline  
 Town 03& 164& 99&60.4\%\\ \hline 
 Overall& 566& 359&63.4\%\\\hline
\end{tabular}
\end{table}

\begin{table}[htp]
\caption{Performance of the software with real-world datasets}
\label{tab:realperf}
\centering
\begin{tabular}{|>{\centering\arraybackslash}p{0.25\linewidth}|>{\centering\arraybackslash}p{0.2\linewidth}|>{\centering\arraybackslash}p{0.2\linewidth}|c|} \hline  
 Dataset& Processed Images &Correct Inference &Accuracy\\ \hline  
Cityscapes stuttgart00&  59& 41&69.5\%\\ \hline  
 Cityscapes stuttgart01&  110& 79&71.8\%\\ \hline  
 Cityscapes stuttgart02& 120& 74&61.7\%\\ \hline 
 nuscenes& 135& 81&60\%\\ \hline 
 nuscenes\_night& 58& 20&34.5\%\\ \hline 
 Overall, without night scenarios& 424& 275&64.9\%\\\hline
 Overall& 482& 295&61.2\%\\\hline
\end{tabular}
\end{table}

With the RGB camera from CARLA, fewer simulations are run since the same process is followed with real-world datasets. In the RGB camera setup, the overall success rate of the software stood at 63.4\% with an average inference time of 0.5s. Due to the semantic segmentation neural network inference, the average time is almost two times from the semantic segmentation camera setup. Compared to the simulation with a semantic segmentation camera, the success rate is reduced drastically with RGB camera simulation. The semantic segmentation algorithm based on DeepLabv3+ with pre-trained weights from the real-world dataset Cityscapes failed at defining boundaries between objects, particularly towards the far end of the frame in the simulation environment. 

\subsubsection{Real-world Portability}\label{porta}
From Table~\ref{tab:realperf}, it can be observed that the software performs best with Cityscapes in classifying the traffic situation with a success rate of about 67\% over three different urban scenarios in Stuttgart. At the same time, the performance drops to 60\% with nuScenes daytime-urban scenarios. With day-time scenarios only (\textit{Overall, without night scenarios} in Table~\ref{tab:realperf}), the software pipeline succeeds in identifying 65\% of the urban traffic situations. The accuracy of situation identification is maximum with the Cityscapes data. Although nuScenes provide high-resolution image data, a drop in performance of the software is observed since \textit{SemSeg} module was pre-trained with Cityscapes data. 

Poor performance in night-time scenarios in nuScenes is expected since semantic segmentation fails without sufficient light. However, a success rate of 34\% is observed at night-time, primarily from correct predictions in well-lit areas of the road where the semantic segmentation network can differentiate between objects. With the night-time scenarios from nuScenes, the software can correctly infer only 34\% of traffic situations. The overall success rate, including the nighttime scenarios in urban areas, is at 61\%. Based on the developed concept, the average inference time of the situation is about 0.37s-0.38s, with about 50\% of the time spent on semantic segmentation. The inference time is lower here than in the simulation prototype, as the CARLA server also requires computing resources to render the simulated world. The vehicles traveling in the opposite direction are also classified as occluding the drive of the ego-vehicle due to the absence of a tracking mechanism. Based on the results, the current software is suitable for real-world applications.

\section{Conclusion} \label{sec:conclusion}
The current work provides a lightweight approach for identifying the overall traffic situation of the ego-vehicle by classifying the semantically segmented BEV image into one of the five defined situation classes using \textit{SIN}. Semantically segmented images bridge the sim2real gap, allowing real-world portability of the concept. Additionally, the use of the \textit{Cam2BEV} method in the software pipeline allows for identifying blind spots or occluded areas around the ego-vehicle, which provides a basis for our concept to be used for situational rule-based applications like relevant data transmission via V2X. This work can provide a foundation for advanced developments in directly identifying vehicle traffic situations.

The main contributions from our work are as follows:
\begin{itemize}
    \item We propose a novel concept for identifying the traffic situation of the vehicle directly using simple and lightweight neural networks.
    \item We verify the concept by setting up a prototype in the CARLA simulation framework \cite{Dosovitskiy17}.
    \item We validate the real-world portability by testing the concept on real-world images from datasets like nuScenes\cite{7780719} and Cityscapes\cite{9156412}.
    \item We provide a basis for selecting relevant data based on the ego-vehicle traffic situation, which can help in redundancy mitigation problems in V2X.
\end{itemize}

As a future outlook on the topic, the proposed concept forms a basis for selecting the relevant sensor data for transmission to other vehicles or selecting relevant data from the backend based on the traffic situation, leading to improved V2X efficiency. The situation identification can be further improved by using tracking algorithms to identify the direction of movement of other traffic participants and by using better neural networks for semantic segmentation. An enhanced dataset with more traffic scenarios will be used to accommodate various sensors like LiDAR or Radar into the situation identification network. This can allow the ego-vehicle to have enhanced situation identification in its vicinity.

\bibliographystyle{IEEEtran}
\bibliography{bib}

\end{document}